\RequirePackage{fix-cm}
\documentclass[twocolumn]{svjour3}          
\usepackage{comment}
\usepackage{amsmath,amsfonts}
\usepackage{subfigure}
\newcommand{\eg}{\emph{e.g.,}~}

\newcommand{\ie}{\emph{i.e.,}~}

\smartqed  
\usepackage[colorlinks,linkcolor=red, citecolor=blue]{hyperref}
\usepackage{graphicx}
\usepackage{cite}
\usepackage{amssymb}
\usepackage[ruled]{algorithm2e} 
\usepackage{url}
\usepackage{multirow}
\usepackage{indentfirst} 
\usepackage{natbib}

\begin{document}

\title{Diff-Font: Diffusion Model for Robust One-Shot Font Generation
}


\author{Haibin He \and Xinyuan Chen \and Chaoyue Wang* \and Juhua Liu* \and Bo Du \and Dacheng Tao \and Qiao Yu 
}


\institute{
Haibin He and Juhua Liu are with the Research Center for Graphic Communication, Printing and Packaging, and Institute of Artificial Intelligence, Wuhan University, Wuhan, China (e-mail: haibinhe@whu.edu.cn; liujuhua@whu.edu.cn).\\
Xinyuan Chen is with the Shanghai AI Laboratory, Shanghai 202150, China (e-mail: xychen9191@gmail.com). \\
Chaoyue Wang is with JD Explore Academy, JD.com, China. (e-mail: chaoyue.wang@outlook.com).\\
Bo Du is with the National Engineering Research Center for Multimedia Software, Institute of Artificial Intelligence, School of Computer Science and Hubei Key Laboratory of Multimedia and Network Communication Engineering, Wuhan University, Wuhan, China (e-mail: dubo@whu.edu.cn).\\
Dacheng Tao is with the School of Computer Science, Faculty of Engineering, The University of Sydney, Australia  (e-mail: dacheng.tao@gmail.com)\\
Yu Qiao is the Shanghai AI Laboratory, Shanghai 202150, China, and also with Shenzhen Institute of Advanced Technology, Chinese Academy of Sciences, Shenzhen 518055, China (e-mail: yu.qiao@siat.ac.cn). \\
Haibin He and Xinyuan Chen contributed equally to this work. Corresponding Authors: Chaoyue Wang (e-mail: chaoyue.wang@outlook.com), Juhua Liu (e-mail: liujuhua@whu.edu.cn).
}

\date{Received: date / Accepted: date}

\maketitle

\begin{abstract}

Font generation presents a significant challenge due to the intricate details needed, especially for languages with complex ideograms and numerous characters, such as Chinese and Korean. Although various few-shot (or even one-shot) font generation methods have been introduced, most of them rely on GAN-based image-to-image translation frameworks that still face (i) unstable training issues, (ii) limited fidelity in replicating font styles, and (iii) imprecise generation of complex characters. To tackle these problems, we propose a unified one-shot font generation framework called Diff-Font, based on the diffusion model. In particular, we approach font generation as a conditional generation task, where the content of characters is managed through predefined embedding tokens and the desired font style is extracted from a one-shot reference image. For glyph-rich characters such as Chinese and Korean, we incorporate additional inputs for strokes or components as fine-grained conditions. Owing to the proposed diffusion training process, these three types of information can be effectively modeled, resulting in stable training. Simultaneously, the integrity of character structures can be learned and preserved. To the best of our knowledge, Diff-Font is the first work to utilize a diffusion model for font generation tasks. Comprehensive experiments demonstrate that Diff-Font outperforms prior font generation methods in both high-fidelity font style replication and the generation of intricate characters. Our method achieves state-of-the-art results in both qualitative and quantitative aspects.

\end{abstract}

\section{Introduction}

\begin{figure*}[t]
  \centering
    \includegraphics[width=16cm]{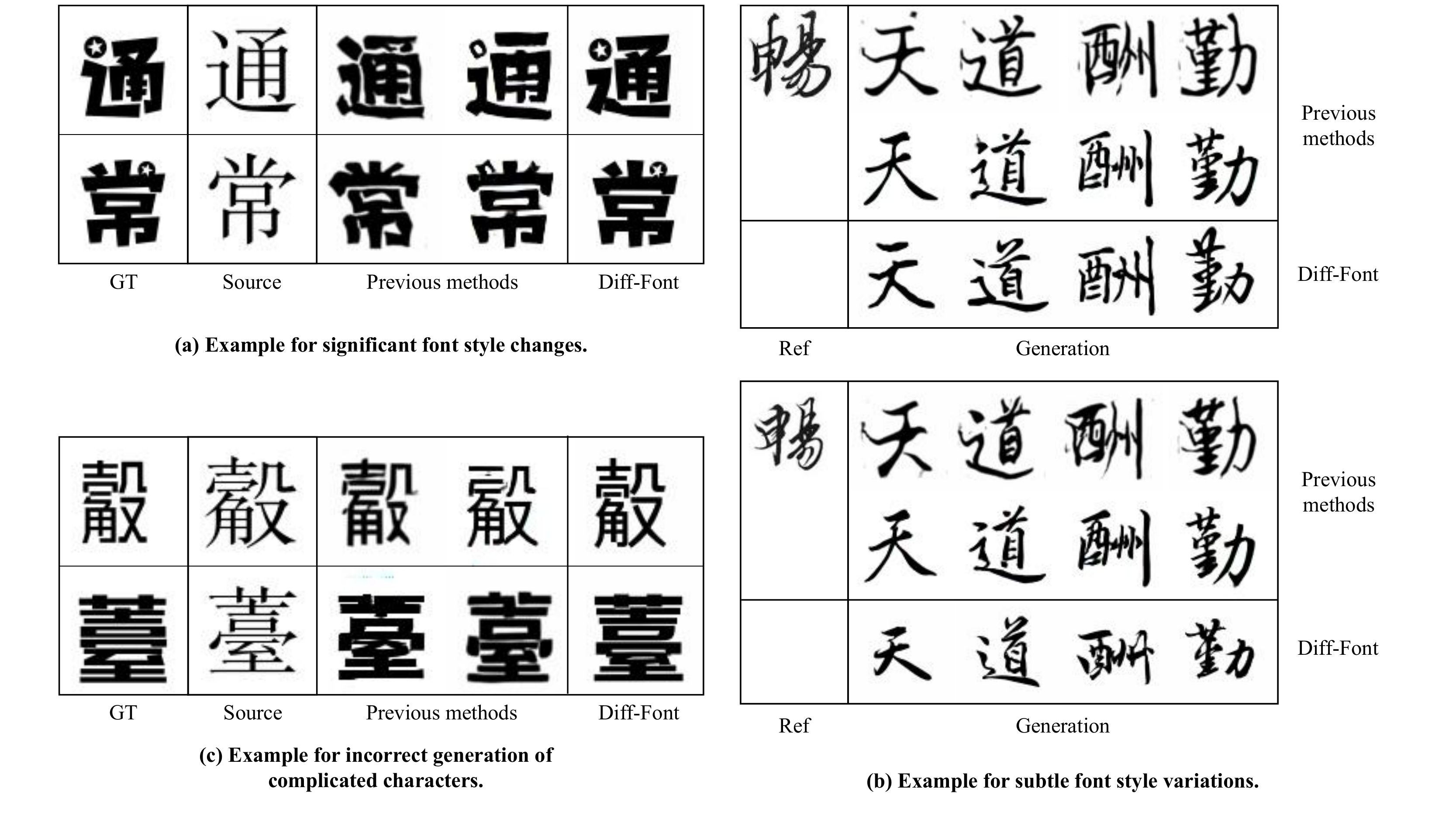}
    \caption{\textbf{Illustration for the problems caused by the gap in font style and complicated characters.} (a) Example of significant font style changes: When the styles between the source and target glyphs differ significantly, methods based on an image-to-image translation framework may generate images with losing local details (column 3 and 4); (b) Example for subtle font style variations: Our proposed Diff-Font can well capture the subtle variations between two fonts with similar styles while previous methods cannot; (c) Example for incorrect generation of complicated character: Image-to-Image translation framework may not perform well in generating characters with complicated structure.}
\label{fig:1}
\end{figure*}

Words are omnipresent in our everyday lives, appearing on book covers, signboards, advertisements, mobile phones, and even clothing. As a result, font generation holds significant commercial value and potential for application. However, designing a font library could be an extremely challenging task, particularly for glyph-rich languages with complex structures, such as Chinese (with over 60,000 glyphs) and Korean (with over 11,000 glyphs). Recently, the progress made in deep generative models, known for their capability to produce high-quality images, has indicated the feasibility of automatically generating diverse font libraries.

``Zi2zi" \cite{Related22} is the first to adopt Generative Adversarial Networks (GANs) \cite{Related44} to automatically generate a Chinese font library by learning a mapping from one style font to another, however, it needs paired data which is labor-reliant and expensive to collect. To facilitate the automatic synthesis of new fonts in an easy manner, numerous Few-shot (or even one-shot) Font Generation (FFG) methods have been proposed. These methods use a character image as the content and a few (or one) target characters to supply the font style, then their models are trained to generate the content character's image with the target font style. Most existing FFG methods are built upon the GAN-based image-to-image translation framework. Some works follow unsupervised methods to obtain content and style features separately, and then fuse them in a generator to generate new characters \cite{Related46}, \cite{Related54}, \cite{Related30}. Meanwhile, some other works exploit auxiliary annotations (\textit{e.g.}, strokes, components) to make the models aware of the specific structure and details about glyphs \cite{Related24},  \cite{Related27}, \cite{Related28}, \cite{Related31}, \cite{Related29}, \cite{Related32}.

Although GAN-based methods have made significant progress and achieved impressive visual quality, font generation remains a notoriously challenging long-tail task due to its stringent demands for intricate details. Most existing methods still grapple with three types of challenges. Firstly, current GAN-based methods, which employ adversarial training schemes, may experience unstable training and convergence difficulties, particularly with large datasets. Secondly, these methods generally treat font generation as a style transfer problem between source and target image domains, often failing to separately model content and font style of characters. Consequently, neither significant font style transfers (\textit{i.e.}, drastic style changes) yield satisfactory results, nor subtle variations between two similar fonts are properly modeled. Last but not the least, when source characters become complex, these methods may struggle to ensure the integrity of the generated character structure. A qualitative illustration of problems arising from gaps in font style and complicated characters can be found in Fig. \ref{fig:1}.

To tackle the aforementioned challenges, we introduce a novel diffusion model-based framework called Diff-Font for one-shot font generation. Instead of treating font generation as a style/domain transfer between a source font domain and a target font domain, the proposed Diff-Font approach considers font generation as a conditional generation task. Specifically, different character content is preprocessed into unique tokens, in contrast to the image inputs employed by previous methods which could cause confusion in similar glyphs.
Regarding font styles, we utilize a pre-trained style encoder to extract style features as our conditional inputs. Moreover, to mitigate imprecise generation issues associated with glyph-rich characters, we incorporate a more fine-grained condition signal to help Diff-Font better model character structures. For Chinese fonts, we use stroke conditions, as strokes represent the smallest units that make up Chinese characters. Likewise, the components of Korean characters serve as the additional conditional input for Korean font generation. Instead of using the one-bit encoding employed in StrokeGAN~\cite{Related25}, we employ count encoding to represent stroke (component) attributes, which more accurately reflects the character's stroke (component) properties. Consequently, the proposed Diff-Font effectively decouples the content and styles of characters, yielding high-quality generation results for complex characters. Simultaneously, thanks to the conditional generation pipeline and diffusion process, Diff-Font can be trained on large-scale datasets while exhibiting improved training stability compared to previous GAN-based methods. Lastly, we assemble a stroke-aware dataset for Chinese font generation and a component-aware dataset for Korean font generation.

In summary, the main contributions of this paper are as follows:
\begin{itemize}
\item  We present Diff-Font, a unified generative network for robust one-shot font generation based on the diffusion model. In comparison to GAN-based methods, Diff-Font offers the advantages of stable training and the ability to be effectively trained on large datasets. To the best of our knowledge, this is the first attempt to develop a diffusion model for font generation.

\item The proposed Diff-Font tackles the font generation task by employing a multi-attribute conditional diffusion model instead of the image-to-image translation framework. Character content and styles are processed as conditions, and the diffusion model utilizes these conditions to generate corresponding character images. Furthermore, a more fine-grained condition, such as stroke or component condition, is employed to enhance the generation of scripts with complex structures. Extensive experiments demonstrate the efficacy of our Diff-Font for one-shot font generation in comparison to previous state-of-the-art methods.

\item We have compiled and annotated a stroke-wise dataset for Chinese and a component-wise dataset for Korean, which we believe can enhance font generation performance from the perspective of strokes and components. The source code, pre-trained models, and datasets are available at \url{https://github.com/Hxyz-123/Font-diff}.

\end{itemize}

The rest of this paper is organized as follows. In Sec.~\ref{sec:Related works}, we briefly review the related works. In Sec.~\ref{sec:Methods}, we introduce our proposed method in detail. Sec.~\ref{sec:Experiments} reports and discusses our experimental results. Lastly, we conclude our study in Sec.~\ref{sec:Conclusion}.

\section{Related Work}
\label{sec:Related works}

\subsection{Image-to-Image Translation}
\label{sec:Image-to-Image Translation}
The task of image-to-image translation involves learning a mapping function that can transform source domain images into corresponding images that preserve the content of the original images while exhibiting the desired style characteristics of the target domain. Generating fonts can be achieved by means of the image-to-image translation models, which can be used to generate any desired font styles from a given content font image. Image-to-image translation using generative adversarial networks (GANs) has been a classical problem in the field of computer vision. Many works have been proposed to address this problem. 
Conditional GAN-based methods \cite{Related02}, such as Pix2Pix \cite{Related01}, require paired data to guide the generation process. To eliminate the dependency on paired data, unsupervised methods have been proposed, including cycle-consistency-based approaches \cite{Related03}, \cite{Related04}, \cite{Related05}, \cite{Related06} and the UNIT \cite{Related11} framework that leverages CoGAN \cite{Related14} and VAE \cite{Related15}. BicycleGAN \cite{Related10} enables one-to-many domain translation by building a bijection between latent coding and output modes. For many-to-many domain translation, methods such as MUNIT \cite{Related08}, CD-GAN \cite{Related09} and FUNIT \cite{Related13} disentangle the content and style representations using two encoders and couple them. Recently, due to the impressive results of the diffusion model, many diffusion model-based methods \cite{Related17}, \cite{Related18}, \cite{Related19}, \cite{Related20}, \cite{Related21} are proposed to tackle image-to-image tasks. However, controlling the generated output using diffusion model-based methods remains a challenge, and further exploration and development are needed, especially in the context of font generation.

Existing image-to-image translation methods generally focus on transforming object pose, texture, color, and style while preserving the content structure, which may not be directly applicable to font generation. Unlike natural images, font styles are primarily defined by variations in shape and specific stroke rules rather than texture and style information. As a result, content structure information may also change during the font generation process. Therefore, applying image-to-image translation methods directly cannot produce satisfactory results.

\subsection{Few-Shot Font Generation}
\label{sec:Few-Shot Font Generation}
Few-shot font generation aims to generate an entire font library with thousands of characters with only a few reference-style images as input. Existing few-shot font generation methods are predominantly based on the image-to-image translation framework, which transfers the source style of content characters to the reference style. To incorporate font-specific prior information into the method or the labels for careful design, various approaches have been proposed, demonstrating the potential of integrating such knowledge to improve the quality and diversity of generated fonts. DG-Font \cite{Related30} implements effective style transfer by replacing the traditional convolutional blocks with deformable convolutional blocks in an unsupervised framework TUNIT \cite{Related33}. ZiGAN \cite{Related23} projects the same character features of different styles into Hilbert space to learn coarse-grained content knowledge. Some methods employ extra information to enhance training, \textit{e.g.}, strokes and components. SC-Font \cite{Related24} uses stroke-level data to improve the correctness of structure and reduce stroke errors in generated images. DM-Font \cite{Related27} employs a dual-memory architecture to disassemble glyphs into stylized components and reassemble them into new glyphs. Its extension version LF-Font \cite{Related28} designs component-wise style encoder and factorization modules to capture local details in rich text design. MX-Font \cite{Related31} has a multi-headed encoder for specializing in different local sub-concepts, such as components, from the given image. 
FS-Font \cite{Related32} proposes a Style Aggregation Module (SAM) and an auxiliary branch to learn the component styles from references and the spatial correspondence between the content and reference glyphs. CG-GAN \cite{Related29} proposes a component discriminator to supervise the generator decoupling content and style at a fine-grained level. 
However, all methods mentioned above are based on GANs, which suffer from instability during training due to their adversarial objective and are prone to mode collapse, leading to suboptimal results especially for font styles with significant or subtle variations. As a result, there remains potential for improvement in the quality of font generation.

\begin{figure*}[t]
  \centering
    \includegraphics[width=16cm]{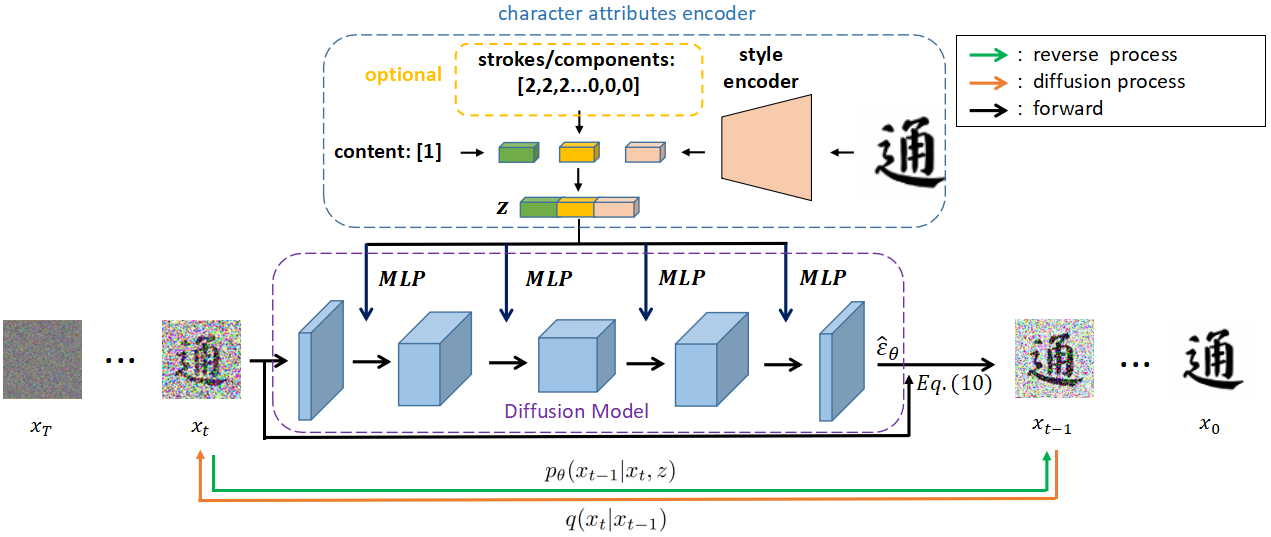}
    \caption{\textbf{Overview of our proposed method}. In the diffusion process, we gradually add noise to image $x_0$, and make it become approximately a Gaussian noise after time step \textit{T}. For the reverse diffusion process, we use a latent variable $z$, which contains the content, style, and other optional attributes semantic information of $x_0$, as a condition to train a diffusion model (based on UNet architecture) to predict the added noise at each time step in the diffusion process.}
\label{fig:2}
\end{figure*}

\subsection{Diffusion Model}
\label{sec:Diffusion Probabilistic Models}
Diffusion Model is a new type of generative model that leverages the iterative reverse diffusion process to generate high-quality images and model complex distributions. It provides state-of-the-art performance in terms of image quality and can generate diverse outputs without mode collapse. Specifically, It employs a Markov chain to convert the Gaussian noise distribution to the real data distribution. Sohl-Dickstein \textit{et al.} \cite{Related34} first clarify the concept of diffusion probabilistic model and denoising diffusion probabilistic models (DDPM) \cite{Related35} improves the theory and proposed to use a UNet to predict the noise added into the image at each diffusion time step. Dhariwal \textit{et al.} \cite{Related36} propose a classifier-guidance mechanism that adopts a pre-trained classifier to provide gradients as guidance toward generating images of the target class. Ho \textit{et al.} \cite{Related37} propose a technique that jointly trains a conditional and an unconditional diffusion model without using a classifier named classifier-free guidance. DDIM \cite{Related47} extends the original DDPM to non-Markovian cases and is able to make accurate predictions with a large step size that reduces the sampling steps to one of the dozens. 
Glide \cite{Related40}, DALL-E2 \cite{Related41}, Imagen \cite{Related42} and Stable Diffusion \cite{Related56} introduce a pre-trained text encoder to generate semantic latent spaces and achieve exceptional results in a text-to-image task. Although the above methods have shown amazing results in image generation, they often focus on generating a specific category of objects or concept-driven generation guided by text prompts, with limited controllability. 

Some other works explore the use of multiple conditions to guide the generation of diffusion models. SDG \cite{Related38} designs a sampling strategy, which adds multi-modal semantic information to the sampling process of the unconditional diffusion model for achieving language guidance and image guidance generation. ILVR \cite{Related39} uses a reference image at each time step during sampling to guide the generation. Diss \cite{Related51} uses stroke images and sketch images as multi-conditions to train a conditional diffusion model to generate images from hand-drawings. Liu \textit{et al.} \cite{Related52} consider the diffusion model as a combination of energy-based models and propose two compositional operators, conjunction and negation, to achieve zero-shot combinatorial generalization to a larger number of objects. Nair \textit{et al.} \cite{Related53} guides the generation of diffusion model by calculating the comprehensive condition scores of multiple modes to solve the problem of multi-modal image generation. ControlNet \cite{Related55} introduces an extra conditional control module to enable a pre-trained diffusion model to be applied to specific tasks. 
This work is further extended by the multi-attribute conditional diffusion model which introduces composite-wise and stroke-wise attributes conditional for better training and attribute-wise diffusion guidance strategy for stroke-aware or component-aware font generation.

\section{Methodology}

\label{sec:method}

\label{sec:Methods}
In this section, we introduce the details of Diff-Font. We first illustrate the framework of our model by incorporating the attributes of content, style, strokes and components (Sec \ref{Sec_3.1}). Then, we elucidate the training process by formulating our multi-attributes conditional diffusion model (Sec \ref{Sec_3.2}). Lastly, we present the adopted strategy to achieve attribute-wise guidance that can set the guidance level of attribute conditions separately during the generation process (Sec \ref{Sec_3.3}).

\subsection{The Framework of Diff-Font}
\label{Sec_3.1}
The framework of our proposed Diff-Font is illustrated in Fig. \ref{fig:2}. As shown, Diff-Font consists of two modules: a character attributes encoder, which encodes the attributes of a character (\ie content, style, strokes, components) into a latent variable, and a diffusion generation model, which uses the latent variable as a condition to generate the character image from Gaussian noise. The character attributes encoder is designed to process the attributes (content, style, strokes, components) of a character image separately. 

In the character attributes encoder $f$, the content (denoted as $c$), style (denoted as $s$), and optional condition (like strokes or components, denoted as $op$) are encoded as the latent variable: $z = f(c, s)$. If using the optional condition, then $z = f(c, s, op)$. 
Unlike previous font generation methods based on image-to-image translation that use the images from the source domain to obtain the content representations, we regard different content characters as different tokens. Similar to word embedding in the NLP community, we adopt an embedding layer to convert different tokens of characters into different content representations. 

\begin{figure}[t]
\centering
  \subfigure[32 basic strokes of Chinese characters.]{
  \includegraphics[width=7.5cm]{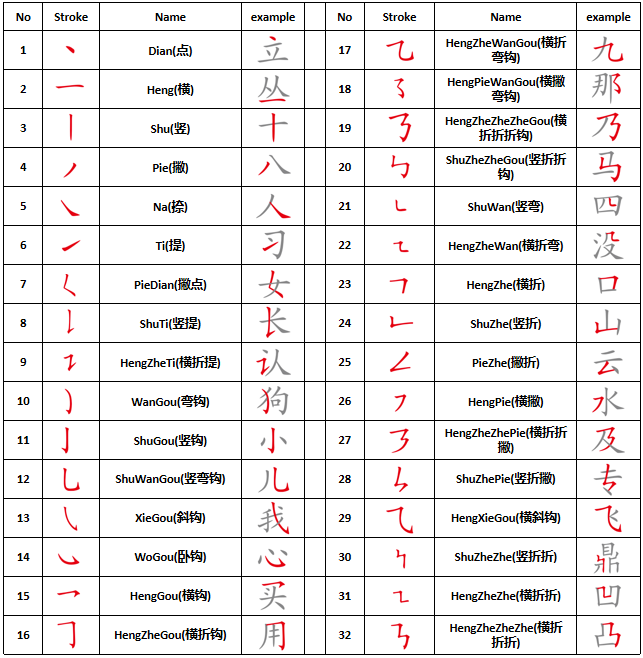}
    }

  \subfigure[Strokes and stroke count encoding vector of Chinese character `Tong'.]{
    \includegraphics[width=7.5cm]{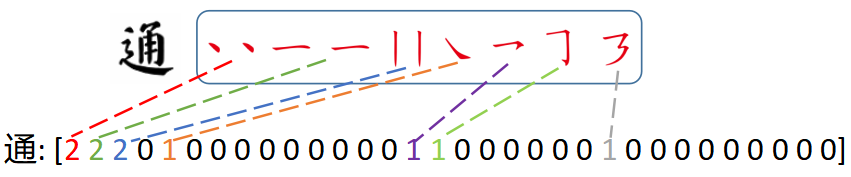}
    }
    \label{fig:3c}
  \caption{(a) 32 basic strokes of Chinese characters. The first and sixth columns are the dimensional locations of the basic strokes in the stroke vector. (b) Strokes and stroke count encoding vector of Chinese character `Tong'. Each dimension of the encoding vector represents the counts of corresponding basic stroke it contains.}
 \label{fig:3}
\end{figure}

\begin{figure}[t]
\centering
  \subfigure[24 basic Korean components.]{
    \includegraphics[width=7cm]{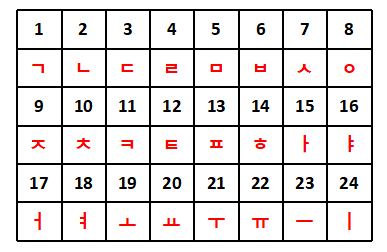}
   }
  \subfigure[Components and count encoding vector of example Korean character.]{
    \includegraphics[width=7.5cm]{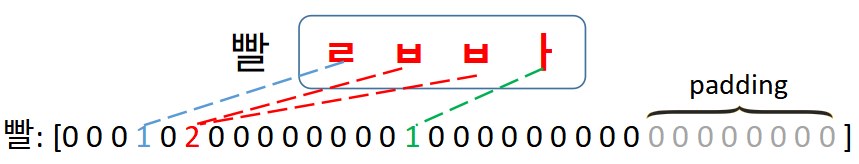}
    }
\caption{(a) 24 basic components of Korean characters. (b) Components and count encoding vector of example Korean character. We encode Korean components in the same way as Chinese strokes. Since Korean has only 24 basic components, we pad into 32 dimensions with 0.}
 \label{fig:4}
\end{figure}

The style representation is extracted by a pre-trained style encoder. A trained style encoder in DG-Font is used as our pre-trained style encoder and its parameters are frozen in our diffusion model training. As for strokes (or components), we encode each character into a 32-dimensional vector. Each dimension of the vector represents the number of corresponding basic strokes (or components) it contains (shown in Fig. \ref{fig:3} and Fig. \ref{fig:4}). This count encoding can better represent the stroke (or component) attribute of a character than one-bit encoding used in StrokeGAN \cite{Related25}. Thereafter, a stroke (or component) vector can be expanded into a vector consistent with the dimension of the content embedding. Using this method, we can obtain attribute representations of a character image and then concatenate them as a condition $z$ for later conditional diffusion model training. 

In the diffusion process, we add random gaussian noise to the real image $x_0$ slowly to obtain a long Markov chain from the real image $x_0$ to noise $x_T$. We adopt UNet architecture as our diffusion model and follow \cite{Related36} to learn the reverse diffusion process. The reverse diffusion process generates characters images from gaussian noise by using multi-attributes condition latent variable $z$. This conditional generation is designed to mitigate the impact of the distinction in font style. 

\subsection{Multi-Attributes Conditional Diffusion Model}
\label{Sec_3.2}

In our method, we regard each raw image of the character which is determined by its content ($c$), style ($s$) (and optional conditions ($op$)) attributes as a sample in the whole training data distribution, and denote the sample as ${x_0 \sim q(x_0 \mid f(c, s))}$. If using the optional condition, then, ${x_0 \sim q(x_0 \mid f(c, s, op))}$. Like the thermal motion of molecules, we add random Gaussian noise to the image thousands of times to gradually transform it from a stable state to a chaotic state. This process is called diffusion process and can be defined as:
\begin{equation}
    q(x_{1:T} \mid x_0) = \prod_{t=1}^{T}q(x_t \mid x_{t-1}),
    \label{eq_1}
\end{equation}
where
\begin{equation}
    q(x_t \mid x_{t-1}) = \mathcal{N}(x_t; \sqrt{1 - \beta_t}x_{t-1}, \beta_t\mathbf{I}),\quad t = 1,...,T, \label{eq_2}
\end{equation}
and $\beta_{1} <  ... < \beta_{T}$ is a variance schedule following \cite{Related35}. According to the Eq.\ref{eq_2}, $x_t$ can be rewritten as:
\begin{align}
x_t &= \sqrt{1 - \beta_{t}}x_{t-1} + \sqrt{\beta_{t}}\epsilon_{t-1} , \quad \epsilon_{t-1} \sim \mathcal{N}(\mathbf{0}, \mathbf{I}) \label{eq_3} \\
&=\sqrt{\bar{\alpha}_t}x_0 + \sqrt{1-\bar{\alpha}_t}\epsilon \label{eq_10}, \bar{\alpha}_t = \prod_{i=1}^{t}\alpha_i,\epsilon \sim \mathcal{N}(\mathbf{0},\mathbf{I}) \\
&\sim \mathcal{N}(x_t; \sqrt{\bar{\alpha}_t}x_0, (1 - \bar{\alpha}_t)\mathbf{I}) \label{eq_11}
\end{align}
where $\alpha_t = 1 - \beta_t$, and $\alpha_t$ is negatively correlated with $\beta_t$, therefore $\alpha_{1} >  ... > \alpha_{T}$. When the $T \xrightarrow[]{} \infty$, $\bar{\alpha}_T$ close to 0, $x_T$ nearly obeys $\mathcal{N}(\mathbf{0},\mathbf{I})$ and the posterior $q(x_{t-1} \mid x_t)$ is also a Gaussian. So in the reverse process, we can sample a noisy image $x_T$ from an isotropic Gaussian and generate the designated character image by denosing the $x_T$ in the long Markov chain with a multi-attributes condition $z = f(c, s)$  (if using the optional condition, then, $z = f(c, s, op)$) that contains the semantic meaning of character. Since the posterior $q(x_{t-1} \mid x_t)$ is hard to estimate, we use $p_\theta$ to approximate the posterior distribution which can be denoted as:
\begin{equation}
p_\theta(x_{0:T} \mid z) = p(x_T)\prod_{t=1}^T p_\theta(x_{t-1} \mid x_{t}, z), \label{eq_12}
\end{equation}
\begin{equation}
p_\theta(x_{t-1} \mid x_{t}, z) = \mathcal{N}(\mu_\theta(x_t, t, z), \Sigma_\theta(x_t, t, z)),  \label{eq_13}
\end{equation}

Following DDPM \cite{Related35}, we set $\Sigma_\theta(x_t, t, z)$ as constants and the diffusion model $\epsilon_{\theta}(x_t, t, z)$ learns to predict the noise $\epsilon$ added to $x_0$ in diffusion process from $x_t$ and condition $z$ for easier training. Through these simplified operations, we can adopt a standard MSE loss to train our multi-attributes-conditional diffusion model:
\begin{equation}
\begin{split}
L_{simple} = \mathbb{E}_{x_0\sim q(x_0), \epsilon\sim \mathcal{N}(\mathbf{0}, \mathbf{I}), z}[\parallel\epsilon - \epsilon_{\theta}(x_t, t, z)\parallel^2].
\end{split}
\end{equation}

\subsection{Attribute-wise Diffusion Guidance Strategy}
\label{Sec_3.3}
 For glyph-rich scripts (\eg Chinese and Korean), we adopt a two-stage training strategy to improve the generation effect. Based on the multi-attributes conditional training (\ie first training stage), we also design a fine-tuning strategy (second training stage) that randomly discards content attribute or stroke (or component) attribute vectors with a $30\%$ probability. If the content and stroke (or component) are discarded at the same time, the style attribute vector also be discarded. Such strategy has two advantages: first, it can enable our model to be more sensitive to these three attributes, and second, it can reduce the number of hyperparameters for we only need two guidance scales instead of three. In our case, we use zero vectors to replace the discarded attribute vectors, denoted as \textbf{0}. When sampling, we modify the predicted noise to $\hat{\epsilon}_\theta$:
\begin{equation}
\begin{split}
&\hat{\epsilon}_\theta(x_t, t, f(c, s, op)) =\epsilon_\theta(x_t,t ,\mathbf{0}) \\
&+ s_1 \ast (\epsilon_\theta(x_t,t,f(c, s, \mathbf{0})) - \epsilon_\theta(x_t,t ,\mathbf{0}))\\
&+ s_2 \ast (\epsilon_\theta(x_t,t,f(\mathbf{0}, s, op)) - \epsilon_\theta(x_t,t ,\mathbf{0})),
\end{split}
\end{equation}
where $s_1$ and $s_2$ are the guidance scales of content and strokes.
Then we adopt DDIM \cite{Related47} to sample on a subset of diffusion steps \{$\tau_1,...,\tau_S$\} and set the variance weight parameter $\eta = 0$ to speed up the generation process. So, we can obtain $x_{\tau_{i-1}}$ from $x_{\tau_i}$ by the following equation:
\begin{equation}
\begin{split}
x_{\tau_{i-1}} = \sqrt{\bar{\alpha}_{\tau_{i-1}}}(\frac{x_{\tau_i} - \sqrt{1 - \bar{\alpha}_{\tau_i}}\hat{\epsilon}_\theta}{\sqrt{\bar{\alpha}_{\tau_i}}}) + \sqrt{1 - \bar{\alpha}_{\tau_{i-1}}}\hat{\epsilon}_\theta.
\end{split}
\end{equation}
The final character image $x_0$ can be obtained by iterating through the above formula.

\section{Experiments}
\label{sec:Experiments}

In this section, we evaluate the performance of the proposed method on the one-shot font generation task by comparing it with state-of-the-art methods. In section ~\ref{sec:Datasets}, we first introduce the datasets and evaluation metrics used to conduct experiments. The implementation details are described in section ~\ref{Implementation Details}.  The results of qualitative and quantitative comparisons between Diff-Font and previous SOTA methods on different script generation are listed in sections ~\ref{sec:Comparison}, ~\ref{sec:Ablation Study}, ~\ref{Korean Script Generation}, ~\ref{Other Script Generation}. Limitations are discussed in section ~\ref{sec:Limitations}.

\subsection{Datasets and Evaluation Metrics}
\label{sec:Datasets}

 \noindent \textbf{Chinese font datasets.} We collect 410 fonts (styles) including handwritten fonts and printed fonts as our whole dataset. Each font has 6,625 Chinese characters that cover almost all commonly used Chinese characters. To evaluate the capacity of methods for different scale datasets, we use a small dataset and a large dataset for experiments. For the small dataset, the training set contains 400 fonts and 800 randomly selected characters, and the testing set contains the remaining 10 fonts with the same characters as the training set. For the large dataset, we use the same 400 fonts but all 6,625 characters in training. The testing set consists of the remaining 10 fonts and 800 characters with complex structures and multiple strokes. In our experiment, the number of small dataset is set consistent with previous methods \cite{Related30}. For fair comparison, the image size is also the same as the previous methods \cite{Related30,Related46}, which is set as 80$\times$80.
 
 \noindent \textbf{Evaluation metrics.} In order to quantitatively compare our method with other advanced methods, we use the common evaluation metrics in image generation task, \textit{e.g.}, SSIM \cite{Related48}, RMSE, LPIPS \cite{Related49}, FID \cite{Related50}. SSIM  (Structural Similarity) imitates the human visual system to compare the structural similarity between two images from three aspects: luminance, contrast and structure. RMSE (Root Mean Square Error) evaluates the similarity between two images by calculating the root mean square error of their pixel values. Both of them are pixel-level metrics. LPIPS (Learned Perceptual Image Patch Similarity), a perceptual-level metric, measures the distance between two images in a deep feature space. FID (Fréchet Inception Distance), measures the difference between generated image and real image in a distribution-wise manner. Moreover, we follow the similar idea in \cite{Related31} to conduct user study for human testing.

\subsection{Implementation Details}
\label{Implementation Details}

\noindent\textbf{Character attributes encoder.} Character attributes encoder in Diff-Font consists of a content embedding layer, a style encoder, a style embedding layer, and an optional embedding layer. The architecture of our style encoder is the same as the style encoder in DG-Font, and the dimensions of the output feature maps are set to 128. Specifically, we adopt an embedding layer for the content attribute and optional attribute respectively, and an MLP for the style attribute. If using the optional attribute, the dimensions of the content, style and optional attribute vectors are set to 128, 128 and 256, respectively. Otherwise, the dimensions of both the content and style vectors are set to 256. Finally, they are concatenated as a 512 dimensions conditional latent vector $z$ for training.

\noindent\textbf{Multi-attributes conditional diffusion model.} Our multi-attributes conditional diffusion model is based on DDPM architecture. We list the hyperparameters setting for our training in TABLE \ref{table_5}. For sampling, we set 25 sampling steps to speed up the generation process.
\begin{table}[h]
\centering
\caption{Hyperparameters setting for multi-attributes conditional diffusion model.}
\setlength{\tabcolsep}{0.2cm}
\footnotesize 
\begin{tabular}{ccc}
\hline
 & Small dataset &  large dataset \\
\hline
Images trained  &320K &2.65M \\
\hline
Batch size &24 &64 \\
\hline
Channels &128 &128 \\
\hline
Res. blocks num &3 &3 \\
\hline
Channel multiplier &1,2,3,4 &1,2,3,4 \\
\hline
Attention resolution & [40,20,10] & [40,20,10] \\
\hline
Diffusion steps &1000 &1000 \\
\hline
Noise Schedule &Linear &Linear \\
\hline
Conditional training iters &300k &420k \\
\hline
fine-tuning iters &300k &380k \\
\hline
Learning rate &1e-4 &1e-4 \\
\hline
Optimizer &\multicolumn{2}{c}{Adam with no weight decay} \\
\hline
Loss &MSE &MSE \\
\hline
\end{tabular}

\label{table_5}
\end{table}
\vspace{-15pt}

\subsection{Comparison with state-of-the-art methods}
\label{sec:Comparison}

In this section, we compare Diff-Font with previous methods for Chinese one-shot font generation: 1) \textbf{FUNIT} \cite{Related13}: FUNIT is a few-shot image-to-image translation framework that disentangles content and style representations by two different encoders and uses AdaIN \cite{Related16} to couple them. 2) \textbf{MX-Font} \cite{Related31}: MX-Font extracts different local sub-concepts by employing multi-headed encoders. 3) \textbf{DG-Font} \cite{Related30}: DG-Font uses the deformable convolution to replace the traditional convolution in an unsupervised framework.  All these methods are based on GANs.

We use both datasets described in Sec \ref{sec:Datasets} to retrain models of FUNIT, MX-Font and DG-Font. During the generation process, only one reference character image with the target font is used. When evaluating these GANs-based methods, we choose the Song font commonly used in the font generation task as the source font \cite{Related30, Related31}.

\noindent \textbf{Quantitative comparison.} TABLE \ref{Table_1} shows the quantitative comparison results between our method and other previous state-of-the-art methods. In the experiments on both small and large datasets, Diff-Font achieves the best performance on all evaluation metrics of SSIM, RMSE, LPIPS and FID. In particular, our method has a great improvement over the second-best method in terms of FID indicators, 22.4$\%$ for the small dataset and 39.2$\%$ for the large dataset. The excellent performance on two scale datasets demonstrates the effectiveness and advantage of our Diff-Font.

\begin{table}[h]
\caption{Quantitative comparison results on two different scale datasets. The best performance is marked in \textbf{bold}.}
\setlength{\tabcolsep}{0.15cm}
\footnotesize 
\centering
\begin{tabular}{ccccc}
\hline
Methods & SSIM($\uparrow$) & RMSE($\downarrow$) & LPIPS($\downarrow$) & FID($\downarrow$) \\
\hline

\multicolumn{5}{c}{Quantitative comparison on small dataset} \\
\hline
FUNIT  &0.700 &0.303 &0.166 &35.20 \\
MX-Font  &0.721 &0.283 &0.151 &37.15 \\
DG-Font  &0.729 &0.28 &0.137 &43.44 \\
Diff-Font(ours) &\textbf{0.742} &\textbf{0.271} &\textbf{0.124} &\textbf{27.30} \\
\hline

\multicolumn{5}{c}{Quantitative comparison on large dataset} \\
\hline
FUNIT  &0.682 &0.311 &0.166 &26.70 \\
MX-Font  &0.692 &0.298 &0.138 &26.64 \\
DG-Font  &0.709 &0.292 &0.112 &28.63 \\
Diff-Font(ours) &\textbf{0.722} &\textbf{0.277} &\textbf{0.104} &\textbf{16.20} \\
\hline
\end{tabular}

\label{Table_1}
\end{table}

\begin{figure*}[ht]
\centering
  \subfigure[Easy styles and easy contents.]{
    \includegraphics[width=15cm]{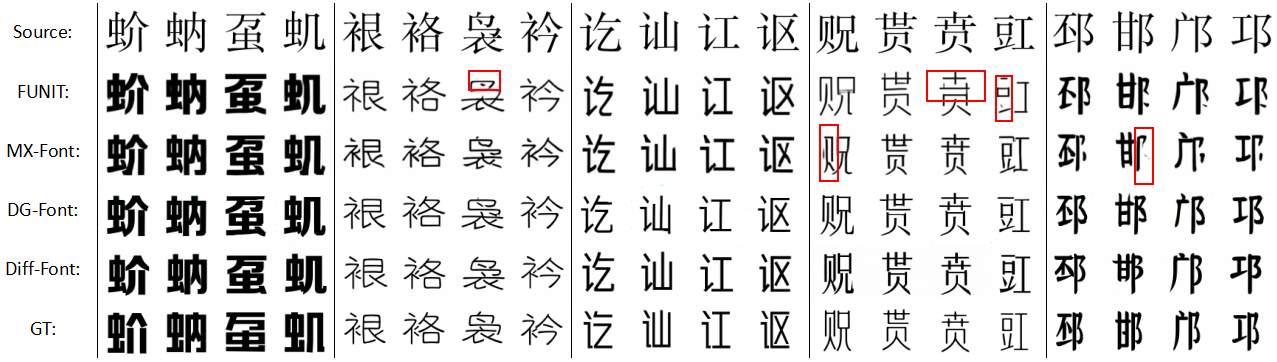}
   }

  \subfigure[Easy styles and difficult contents.]{
    \includegraphics[width=15cm]{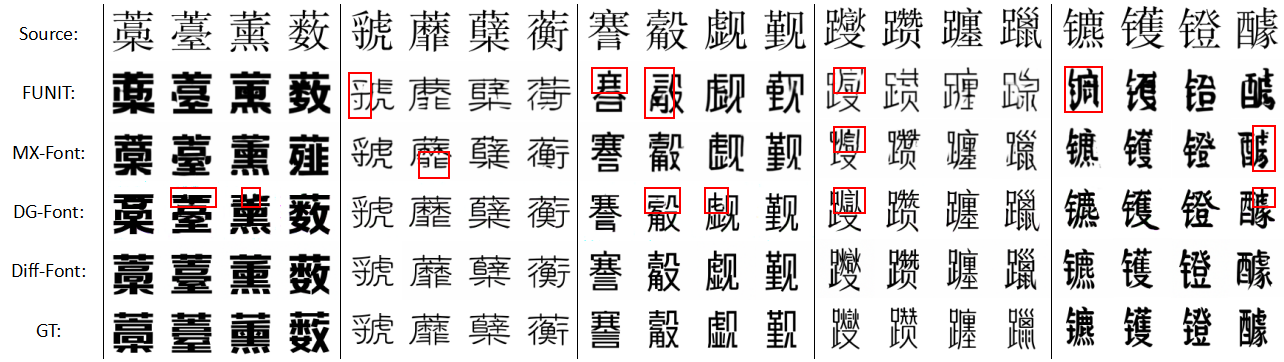}
    }

  \subfigure[Difficult styles and difficult contents.]{
    \includegraphics[width=15cm]{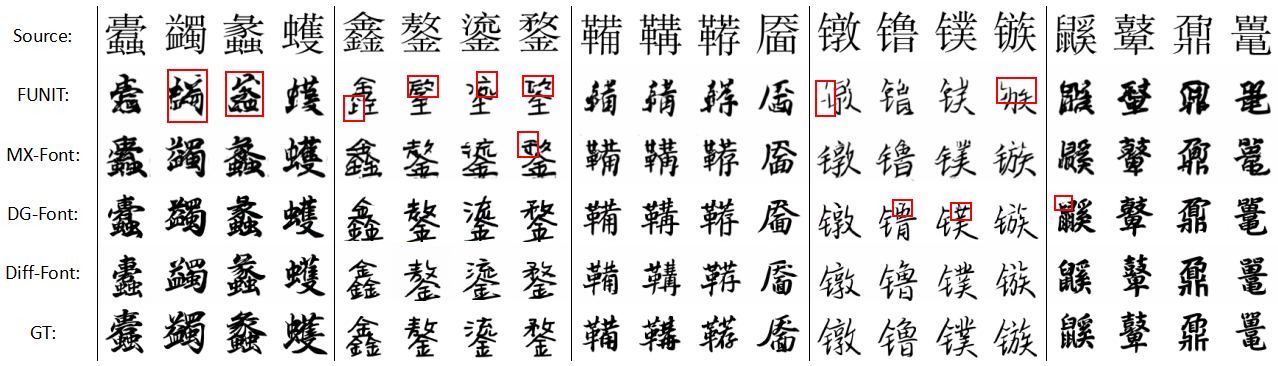}
    }
  \caption{\textbf{Example generation results on large test dataset}. Easy style means the style of the reference font is similar to the source font. The characters with 10 or fewer strokes are easy contents, and those with 15 or more are difficult contents.}
 \label{fig:5}
\end{figure*}

\begin{figure*}[ht]
  \centering
    \includegraphics[width=15cm]{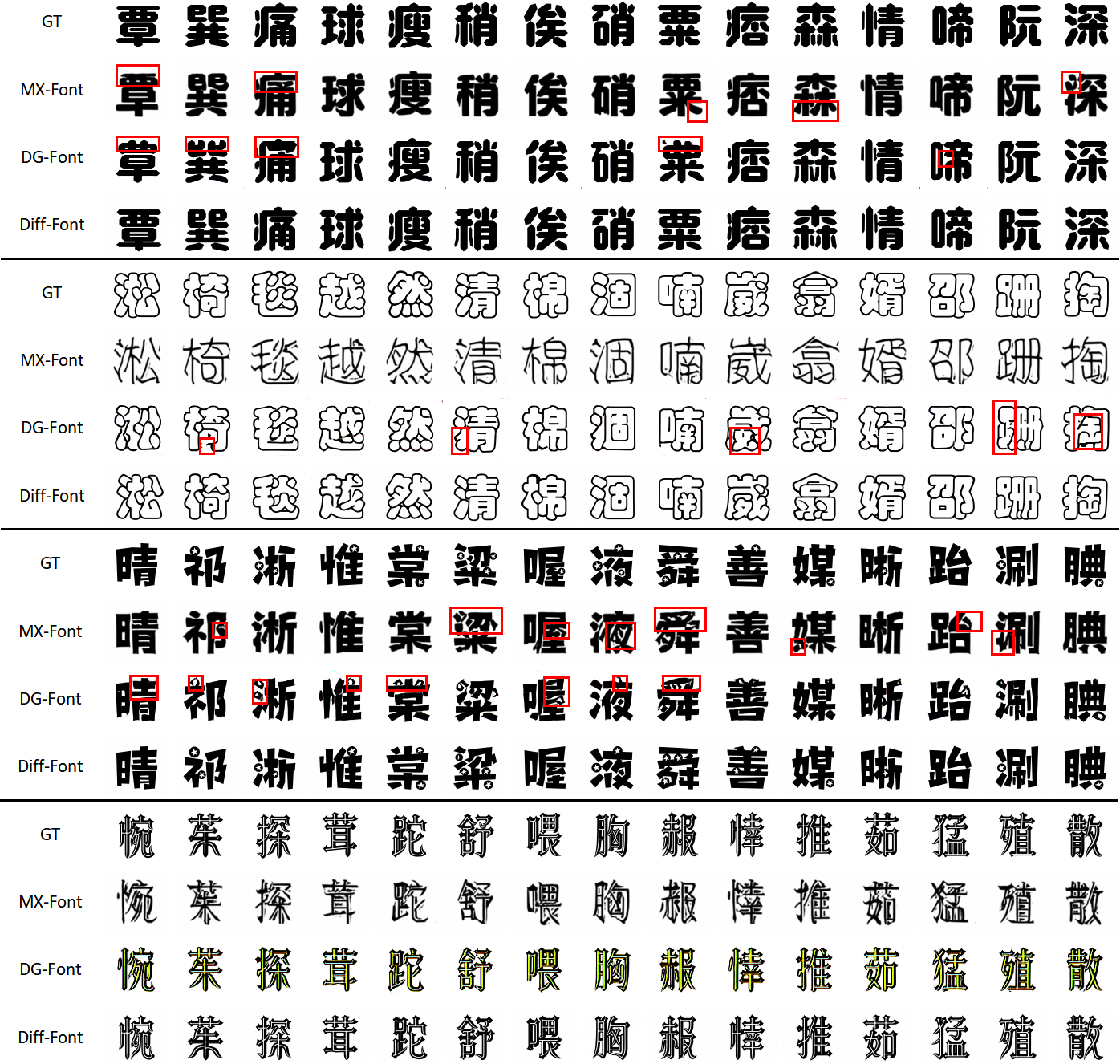}
    \caption{Example generation results of MX-Font, DG-Font, Diff-Font on four art fonts. It can be seen that the structure of the characters generated by MX-Font is severely distorted and the characters generated by DG-Font may contain artifacts.}
\label{fig:6}
\end{figure*}

\noindent \textbf{Qualitative comparison.} The qualitative comparison results are shown in Fig. \ref{fig:5}. For qualitative comparison, we define style and content based on the difficulty of implementation as follows. The target styles similar to the source font are regarded as easy styles, otherwise as difficult styles. The characters with the number of strokes less than or equal to 10 are defined as easy contents, and the characters with the number of strokes more than or equal to 15 as difficult contents. We make qualitative comparisons under the three settings of ESEC (easy styles and easy contents), ESDC (easy styles and difficult contents), and DSDC (difficult styles and difficult contents), respectively. As shown in Fig. \ref{fig:5}, FUNIT often generates incomplete characters, and when the character structure is more complex, it would produce distorted structures. MX-Font could maintain the shape of characters to a certain extent, but it tends to generate vague characters and unclear backgrounds. DG-Font performs well in ESEC task, but losses some important stroke detailed local components in ESDC and DSDC tasks. Compared to these previous methods, our proposed Diff-Font could generate high quality character images in all three tasks. 
\begin{figure}[t]
\centering
    \includegraphics[width=7.5cm]{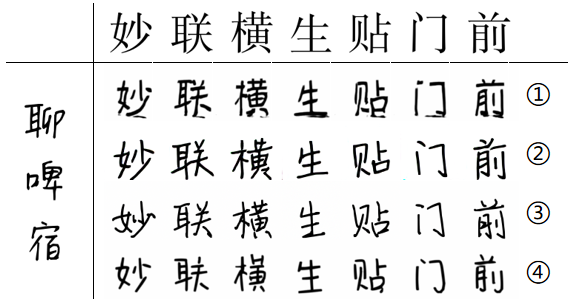}
\caption{\textbf{An example for human testing}. The first column shows three characters with the reference target style, and the first row lists characters with source content.}
\label{fig:7}
\end{figure}

In addition, Fig. \ref{fig:6} shows more qualitative comparison results on four chosen art fonts to better illustrate the effectiveness and advantages of Diff-Font. As these comparison results, when there is significant stylistic difference between the source and target font, GAN-based image-to-image translation frameworks would lead to worse structural distortion and loss of details, and our proposed Diff-Font based on conditional diffusion model could effectively reduce the occurrence.

 \noindent \textbf{Human testing.} We conducted a user study with 10 test fonts, as specified in Sec \ref{sec:Datasets}. Each method was applied to generate a line of ancient Chinese poetry on each font, and 64 participants were asked to evaluate the results based on both style and content. Participants chosen their favorite output, so we obtained 64$\times$10 = 640 results and calculated the percentage of scores for each method. The visualization of generation example is shown in Fig. \ref{fig:7}, and study results are presented in TABLE \ref{table:human testing}. As can be seen, our proposed Diff-Font achieves the best score in human testing, which also verifies the effectiveness of our proposed framework.
\begin{table}[h]
\setlength{\tabcolsep}{0.4cm}
\caption{Results of Human testing.Best results is marked in \textbf{bold}.}
\centering
\footnotesize 
\begin{tabular}{cccc}
\hline
FUNIT & MX-Font & DG-Font & Diff-Font \\
\hline
13.44\%  & 13.28\% & 24.53\% &  \textbf{48.75\%} \\
\hline
\label{table:human testing}
\end{tabular}
\end{table}


\subsection{Ablation Studies}
\label{sec:Ablation Study}
In this part, we further conduct ablation studies to evaluate the effectiveness of the stroke count encoding, and discuss the impact of guidance scales.

\begin{figure}[t]
    \centering
    \includegraphics[width=8cm]{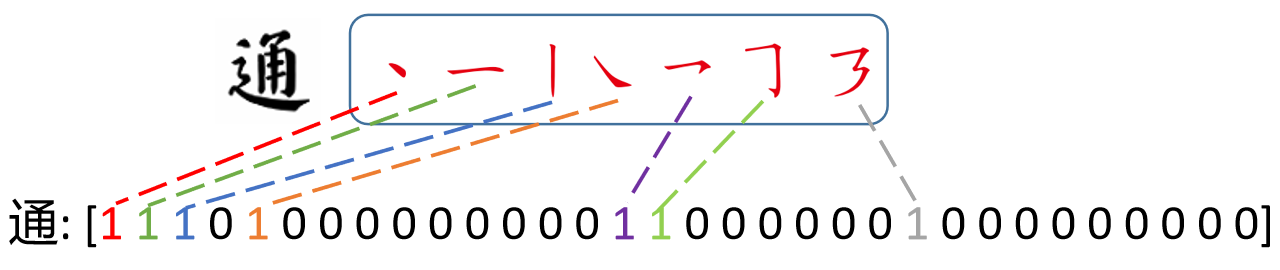}
    \caption{\textbf{One-bit stroke encoding in StrokeGAN\cite{Related25}.} Each dimension of the encoding vector indicates whether the character contains the corresponding basic stroke.}
    \label{fig:8}
\end{figure}

\begin{figure}[t]
  \centering
    \includegraphics[width=8cm]{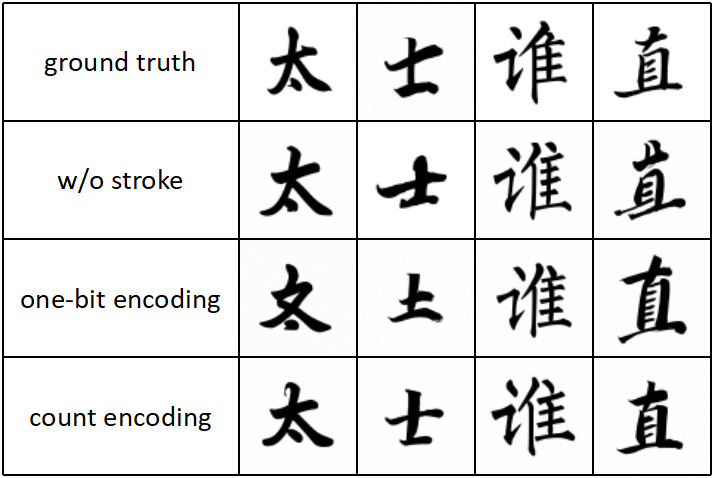}
\caption{\textbf{Qualitative results of ablation studies using different stroke condition.}  The first row is the ground truth, and from the second to the fourth row are results of Diff-Font without stroke condition, with one-bit stroke encoding, with stroke count encoding, respectively. 
    }
\label{fig:9}
\end{figure}

\noindent \textbf{Effectiveness of the stroke count encoding.}
We train three Diff-Font separately on the small dataset, one does not use the stroke condition, one uses the one-bit encoding stroke condition and the remaining one uses the count encoding stroke condition. As is shown in TABLE \ref{table_3}, using count encoding stroke condition achieves the best quantitative results in all evaluation metrics among the three models and we can observe that adding the one-bit encoding stroke condition(Fig. \ref{fig:8}) even causes a decline in model performance. In the visualization result of columns 2 and 3 in Fig. \ref{fig:9}, we find that other characters with the same basic strokes are generated when using the one-bit encoding. And according to column 4 and column 5 in Fig. \ref{fig:9}, when in the case of generating a difficult structure character, Diff-Font without stroke condition and Diff-Font with one-bit encoding may generate characters with stroke errors since the number of basic strokes is not explicitly encoded. These reveals that count encoding is effective for improving the quality by preserving a completed number of strokes.
\begin{table}[h]
\caption{Effectiveness of the stroke count encoding form \\ versus one-bit stroke encoding.Best results is marked in \textbf{bold}.}
\setlength{\tabcolsep}{0.10cm}
\centering
\footnotesize 
\begin{tabular}{ccccc}
\hline
Methods & SSIM($\uparrow$) & RMSE($\downarrow$) & LPIPS($\downarrow$) & FID($\downarrow$) \\
\hline
w/o strokes  &0.74 &0.275 &0.127 &28.83 \\
one-bit encoding  &0.739 &0.277 &0.131 &30.44 \\
count encoding &\textbf{0.742} &\textbf{0.271} &\textbf{0.124} &\textbf{27.30} \\
\hline
\end{tabular}
\label{table_3}
\end{table}

\noindent \textbf{Impact of guidance scales.}
We further discuss the impact of content and stroke on the generation by setting different content scales ($s_1$) and stroke scales ($s_2$). Our experiments are conducted on the test set in large dataset mentioned in Sec \ref{sec:Datasets}. In TABLE \ref{table_scale}, we obtain that using the setting $s_1$ = 3, $s_2$ = 3 can get the best quality generated images.

\begin{table}[h]
\caption{Impact of guidance scales. The best and second-best result are marked in \textbf{bold} and \underline{underlined}, respectively.}
\setlength{\tabcolsep}{0.15cm}
\centering
\footnotesize 
\begin{tabular}{ccccc}
\hline
Scales & SSIM($\uparrow$) & RMSE($\downarrow$) & LPIPS($\downarrow$) & FID($\downarrow$) \\
\hline
$s_1 = 1, s_2 = 1$  &0.720 &0.280 &0.108 &16.67 \\
$s_1 = 1, s_2 = 3$  &0.720 &0.281 &0.112 &16.88 \\
$s_1 = 1, s_2 = 5$  &0.716 &0.285 &0.120 &17.16 \\
$s_1 = 3, s_2 = 1$  &\textbf{0.722} &0.279 &0.105 &16.36 \\
$s_1 = 3, s_2 = 3$  &\textbf{0.722} &\textbf{0.277} &\textbf{0.104} &\underline{16.20} \\
$s_1 = 5, s_2 = 1$  &0.720 &0.280 &0.107 &\textbf{16.18} \\
$s_1 = 5, s_2 = 3$  &0.721 &\underline{0.278} &\textbf{0.104} &16.27 \\
\hline
\end{tabular}
\label{table_scale}
\end{table}

\subsection{Korean Script Generation}
\label{Korean Script Generation}

Our proposed Diff-Font is language independent, so it provides potential general solution for font generation in different languages by utilizing various attribute conditions. In this section, we evaluate the effectiveness of Diff-Font in Korean. As illustrated in Fig. \ref{fig:4}, the Chinese stroke condition can be substituted with the component condition of Korean.

\begin{figure}[t]
\centering
    \includegraphics[width=8cm]{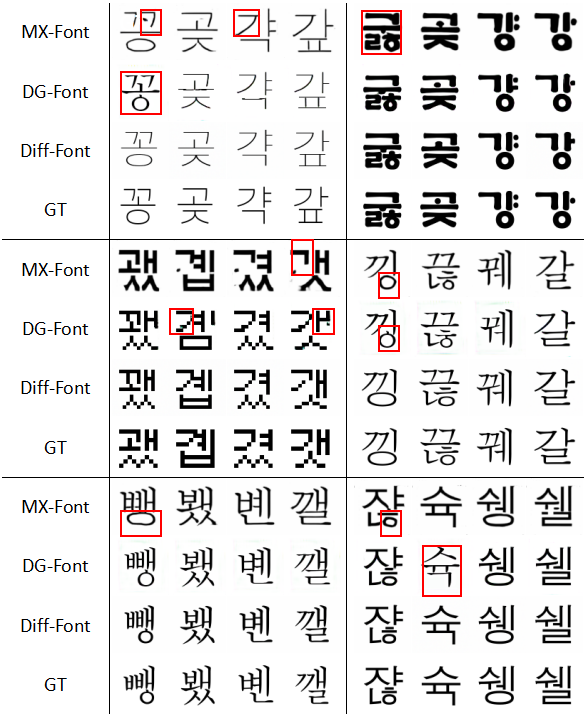}
\caption{Qualitative results on Korean script}
\label{fig:10}
\end{figure}

\begin{table}[t]
\caption{Quantitative results on Korean script.Best results is marked in \textbf{bold}.}
\setlength{\tabcolsep}{0.15cm}
\centering
\footnotesize 
\begin{tabular}{ccccc}
\hline
Methods & SSIM($\uparrow$) & RMSE($\downarrow$) & LPIPS($\downarrow$)  & FID($\downarrow$) \\
\hline
MX-Font  & 0.691 & 0.278  & 0.158 & 47.05 \\
DG-Font  & 0.771 & 0.235 & 0.095& 43.36\\
Diff-Font &\textbf{0.812} &\textbf{0.196} &\textbf{0.072} &\textbf{10.69} \\
\hline
\label{table:gen_korean}
\end{tabular}
\end{table}

Specifically, we collect a dataset of 201 Korean fonts, 195 for training, and the remaining 6 for testing. This dataset contains 2,350 Korean characters.
To evaluate the effectiveness of our proposed method, we conducted comparisons with the DG-Font and MX-Font approaches in generating 800 Korean characters and the results are presented in TABLE \ref{table:gen_korean} and Fig. \ref{fig:10}. We can see that our method also achieves the best results in generating Korean script.

\subsection{Other Script Generation}
\label{Other Script Generation}
\begin{figure}[h]
  \centering
    \includegraphics[width=8cm]{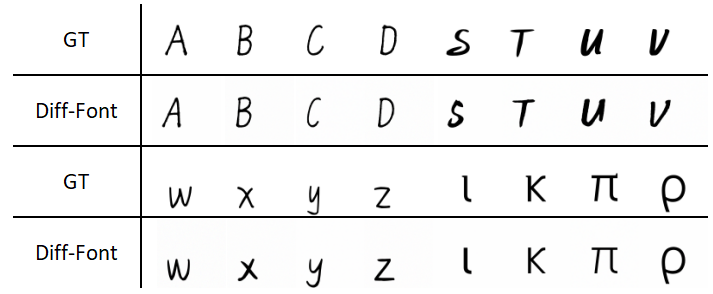}
    \caption{Example generation results of Diff-Font on Latin and Greek.}
\label{fig:11}
\end{figure}

As for some simple scripts without complex structures (\eg Latin and Greek), we can train a Diff-Font in the first stage by only using content and style attribute conditions without fine-tuning in the second stage. As shown in Fig. \ref{fig:11}, our model is also effective in Latin and Greek font generation.

\subsection{Limitations}
\label{sec:Limitations}
\begin{figure}[h]
  \centering
    \includegraphics[width=8cm]{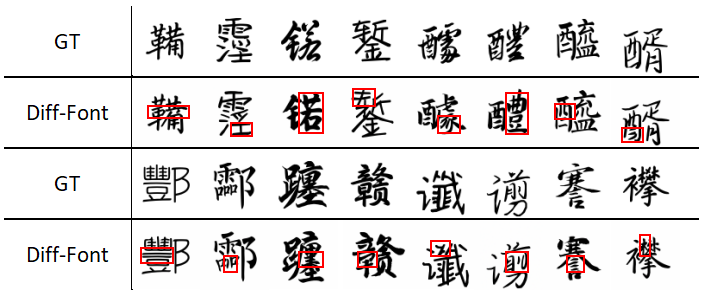}
    \caption{Some failure cases. Characters with extreme complex structures or uncommon styles still suffer generation failures.}
\label{fig:12}
\end{figure}

As our proposed Diff-Font is based on the denoising diffusion model, it has the same problem as most existing diffusion models with low inference efficiency. Moreover, our experimental results show that equipping with stroke/component condition for font generation could reduce generation errors, but cannot completely eliminate them. Some characters with extreme intricate structures or uncommon styles that were infrequently encountered in the training set still suffer generation failures. Some failure cases are shown in Fig. \ref{fig:12}.

\section{Conclusion}
\label{sec:Conclusion}
In this paper, we propose a unified method based on the diffusion model, namely Diff-Font, for one-shot font generation task. The proposed Diff-Font has a stable training process and can be well-trained on large datasets. To address the problems of unsatisfactory generation results on large or subtle differences in the style of source font and target font faced by previous GANs-based methods, we regard font generation as a conditional generation task and generate the corresponding character images according to the given character attribute conditions. Furthermore, we introduce stroke- and component-wise information to improve the structural integrity of generated characters and solve the problem of low generation quality of complicated characters for Chinese and Korean generation. The remarkable performance on two datasets with different scales shows the effectiveness of Diff-Font.

\begin{acknowledgements}
This work was supported in part by the National Natural Science Foundation of China under Grants 62076186, 62225113, and 62102150, and in part by the Science and Technology Major Project of Hubei Province (Next-Generation AI Technologies) under Grant 2019AEA170. The numerical calculations in this paper have been done on the supercomputing system in the Supercomputing Center of Wuhan University.
\end{acknowledgements}

%
%

\bibliographystyle{spbasic}      
\bibliography{reference.bib}   

\begin{thebibliography}{51}
\providecommand{\natexlab}[1]{#1}
\providecommand{\url}[1]{{#1}}
\providecommand{\urlprefix}{URL }
\expandafter\ifx\csname urlstyle\endcsname\relax
  \providecommand{\doi}[1]{DOI~\discretionary{}{}{}#1}\else
  \providecommand{\doi}{DOI~\discretionary{}{}{}\begingroup
  \urlstyle{rm}\Url}\fi
\providecommand{\eprint}[2][]{\url{#2}}

\bibitem[{An and Cho(2015)}]{Related15}
An J, Cho S (2015) Variational autoencoder based anomaly detection using
  reconstruction probability. Special Lecture on IE 2(1):1--18

\bibitem[{Baek et~al.(2021)Baek, Choi, Uh, Yoo, and Shim}]{Related33}
Baek K, Choi Y, Uh Y, Yoo J, Shim H (2021) Rethinking the truly unsupervised
  image-to-image translation. In: Proceedings of the IEEE/CVF International
  Conference on Computer Vision, pp 14154--14163

\bibitem[{Cha et~al.(2020)Cha, Chun, Lee, Lee, Kim, and Lee}]{Related27}
Cha J, Chun S, Lee G, Lee B, Kim S, Lee H (2020) Few-shot compositional font
  generation with dual memory. In: European Conference on Computer Vision,
  Springer, pp 735--751

\bibitem[{Cheng et~al.(2022)Cheng, Chen, Chiu, Lee, and Tseng}]{Related51}
Cheng SI, Chen YJ, Chiu WC, Lee HY, Tseng HY (2022) Adaptively-realistic image
  generation from stroke and sketch with diffusion model. arXiv e-prints pp
  arXiv--2208

\bibitem[{Choi et~al.(2021)Choi, Kim, Jeong, Gwon, and Yoon}]{Related39}
Choi J, Kim S, Jeong Y, Gwon Y, Yoon S (2021) Ilvr: Conditioning method for
  denoising diffusion probabilistic models. arXiv preprint arXiv:210802938

\bibitem[{Dhariwal and Nichol(2021)}]{Related36}
Dhariwal P, Nichol A (2021) Diffusion models beat gans on image synthesis.
  Advances in Neural Information Processing Systems 34:8780--8794

\bibitem[{Gao et~al.(2019)Gao, Guo, Lian, Tang, and Xiao}]{Related54}
Gao Y, Guo Y, Lian Z, Tang Y, Xiao J (2019) Artistic glyph image synthesis via
  one-stage few-shot learning. ACM Transactions on Graphics (TOG) 38(6):1--12

\bibitem[{Goodfellow et~al.(2020)Goodfellow, Pouget-Abadie, Mirza, Xu,
  Warde-Farley, Ozair, Courville, and Bengio}]{Related44}
Goodfellow I, Pouget-Abadie J, Mirza M, Xu B, Warde-Farley D, Ozair S,
  Courville A, Bengio Y (2020) Generative adversarial networks. Communications
  of the ACM 63(11):139--144

\bibitem[{Heusel et~al.(2017)Heusel, Ramsauer, Unterthiner, Nessler, and
  Hochreiter}]{Related50}
Heusel M, Ramsauer H, Unterthiner T, Nessler B, Hochreiter S (2017) Gans
  trained by a two time-scale update rule converge to a local nash equilibrium.
  Advances in neural information processing systems 30

\bibitem[{Ho and Salimans(2022)}]{Related37}
Ho J, Salimans T (2022) Classifier-free diffusion guidance. arXiv preprint
  arXiv:220712598

\bibitem[{Ho et~al.(2020)Ho, Jain, and Abbeel}]{Related35}
Ho J, Jain A, Abbeel P (2020) Denoising diffusion probabilistic models.
  Advances in Neural Information Processing Systems 33:6840--6851

\bibitem[{Huang and Belongie(2017)}]{Related16}
Huang X, Belongie S (2017) Arbitrary style transfer in real-time with adaptive
  instance normalization. In: Proceedings of the IEEE international conference
  on computer vision, pp 1501--1510

\bibitem[{Huang et~al.(2018)Huang, Liu, Belongie, and Kautz}]{Related08}
Huang X, Liu MY, Belongie S, Kautz J (2018) Multimodal unsupervised
  image-to-image translation. In: Proceedings of the European conference on
  computer vision (ECCV), pp 172--189

\bibitem[{Isola et~al.(2017)Isola, Zhu, Zhou, and Efros}]{Related01}
Isola P, Zhu JY, Zhou T, Efros AA (2017) Image-to-image translation with
  conditional adversarial networks. In: Proceedings of the IEEE conference on
  computer vision and pattern recognition, pp 1125--1134

\bibitem[{Jiang et~al.(2019)Jiang, Lian, Tang, and Xiao}]{Related24}
Jiang Y, Lian Z, Tang Y, Xiao J (2019) Scfont: Structure-guided chinese font
  generation via deep stacked networks. In: Proceedings of the AAAI conference
  on artificial intelligence, pp 4015--4022

\bibitem[{Kancharagunta and Dubey(2019)}]{Related06}
Kancharagunta KB, Dubey SR (2019) Csgan: Cyclic-synthesized generative
  adversarial networks for image-to-image transformation. arXiv preprint
  arXiv:190103554

\bibitem[{Kim et~al.(2017)Kim, Cha, Kim, Lee, and Kim}]{Related05}
Kim T, Cha M, Kim H, Lee JK, Kim J (2017) Learning to discover cross-domain
  relations with generative adversarial networks. In: International conference
  on machine learning, PMLR, pp 1857--1865

\bibitem[{Kong et~al.(2022)Kong, Luo, Ma, Zhu, Zhu, Yuan, and Jin}]{Related29}
Kong Y, Luo C, Ma W, Zhu Q, Zhu S, Yuan N, Jin L (2022) Look closer to
  supervise better: One-shot font generation via component-based discriminator.
  In: Proceedings of the IEEE/CVF Conference on Computer Vision and Pattern
  Recognition, pp 13482--13491

\bibitem[{Li et~al.(2022)Li, Xue, Liu, and Lai}]{Related20}
Li B, Xue K, Liu B, Lai YK (2022) Vqbb: Image-to-image translation with vector
  quantized brownian bridge. arXiv preprint arXiv:220507680

\bibitem[{Liu and Tuzel(2016)}]{Related14}
Liu MY, Tuzel O (2016) Coupled generative adversarial networks. Advances in
  neural information processing systems 29

\bibitem[{Liu et~al.(2017)Liu, Breuel, and Kautz}]{Related11}
Liu MY, Breuel T, Kautz J (2017) Unsupervised image-to-image translation
  networks. Advances in neural information processing systems 30

\bibitem[{Liu et~al.(2019)Liu, Huang, Mallya, Karras, Aila, Lehtinen, and
  Kautz}]{Related13}
Liu MY, Huang X, Mallya A, Karras T, Aila T, Lehtinen J, Kautz J (2019)
  Few-shot unsupervised image-to-image translation. In: Proceedings of the
  IEEE/CVF international conference on computer vision, pp 10551--10560

\bibitem[{Liu et~al.(2022)Liu, Li, Du, Torralba, and Tenenbaum}]{Related52}
Liu N, Li S, Du Y, Torralba A, Tenenbaum JB (2022) Compositional visual
  generation with composable diffusion models. arXiv preprint arXiv:220601714

\bibitem[{Liu et~al.(2021)Liu, Park, Azadi, Zhang, Chopikyan, Hu, Shi,
  Rohrbach, and Darrell}]{Related38}
Liu X, Park DH, Azadi S, Zhang G, Chopikyan A, Hu Y, Shi H, Rohrbach A, Darrell
  T (2021) More control for free! image synthesis with semantic diffusion
  guidance. arXiv preprint arXiv:211205744

\bibitem[{Mirza and Osindero(2014)}]{Related02}
Mirza M, Osindero S (2014) Conditional generative adversarial nets. CoRR

\bibitem[{Nair et~al.(2022)Nair, Bandara, and Patel}]{Related53}
Nair NG, Bandara WGC, Patel VM (2022) Image generation with multimodal priors
  using denoising diffusion probabilistic models. arXiv preprint
  arXiv:220605039

\bibitem[{Nichol et~al.(2021)Nichol, Dhariwal, Ramesh, Shyam, Mishkin, McGrew,
  Sutskever, and Chen}]{Related40}
Nichol A, Dhariwal P, Ramesh A, Shyam P, Mishkin P, McGrew B, Sutskever I, Chen
  M (2021) Glide: Towards photorealistic image generation and editing with
  text-guided diffusion models. arXiv preprint arXiv:211210741

\bibitem[{Park et~al.(2021)Park, Chun, Cha, Lee, and Shim}]{Related31}
Park S, Chun S, Cha J, Lee B, Shim H (2021) Multiple heads are better than one:
  Few-shot font generation with multiple localized experts. In: Proceedings of
  the IEEE/CVF International Conference on Computer Vision, pp 13900--13909

\bibitem[{Park et~al.(2022)Park, Chun, Cha, Lee, and Shim}]{Related28}
Park S, Chun S, Cha J, Lee B, Shim H (2022) Few-shot font generation with
  weakly supervised localized representations. IEEE Transactions on Pattern
  Analysis and Machine Intelligence

\bibitem[{Ramesh et~al.(2022)Ramesh, Dhariwal, Nichol, Chu, and
  Chen}]{Related41}
Ramesh A, Dhariwal P, Nichol A, Chu C, Chen M (2022) Hierarchical
  text-conditional image generation with clip latents. arXiv preprint
  arXiv:220406125

\bibitem[{Rombach et~al.(2022)Rombach, Blattmann, Lorenz, Esser, and
  Ommer}]{Related56}
Rombach R, Blattmann A, Lorenz D, Esser P, Ommer B (2022) High-resolution image
  synthesis with latent diffusion models. In: Proceedings of the IEEE/CVF
  Conference on Computer Vision and Pattern Recognition, pp 10684--10695

\bibitem[{Saharia et~al.(2022{\natexlab{a}})Saharia, Chan, Chang, Lee, Ho,
  Salimans, Fleet, and Norouzi}]{Related17}
Saharia C, Chan W, Chang H, Lee C, Ho J, Salimans T, Fleet D, Norouzi M
  (2022{\natexlab{a}}) Palette: Image-to-image diffusion models. In: ACM
  SIGGRAPH 2022 Conference Proceedings, pp 1--10

\bibitem[{Saharia et~al.(2022{\natexlab{b}})Saharia, Chan, Saxena, Li, Whang,
  Denton, Ghasemipour, Ayan, Mahdavi, Lopes et~al.}]{Related42}
Saharia C, Chan W, Saxena S, Li L, Whang J, Denton E, Ghasemipour SKS, Ayan BK,
  Mahdavi SS, Lopes RG, et~al. (2022{\natexlab{b}}) Photorealistic
  text-to-image diffusion models with deep language understanding. arXiv
  preprint arXiv:220511487

\bibitem[{Sasaki et~al.(2021)Sasaki, Willcocks, and Breckon}]{Related18}
Sasaki H, Willcocks CG, Breckon TP (2021) Unit-ddpm: Unpaired image translation
  with denoising diffusion probabilistic models. arXiv preprint arXiv:210405358

\bibitem[{Sohl-Dickstein et~al.(2015)Sohl-Dickstein, Weiss, Maheswaranathan,
  and Ganguli}]{Related34}
Sohl-Dickstein J, Weiss E, Maheswaranathan N, Ganguli S (2015) Deep
  unsupervised learning using nonequilibrium thermodynamics. In: International
  Conference on Machine Learning, PMLR, pp 2256--2265

\bibitem[{Song et~al.(2020)Song, Meng, and Ermon}]{Related47}
Song J, Meng C, Ermon S (2020) Denoising diffusion implicit models. arXiv
  e-prints pp arXiv--2010

\bibitem[{Tang et~al.(2022)Tang, Cai, Liu, Hong, Gong, Fan, Han, Liu, Ding, and
  Wang}]{Related32}
Tang L, Cai Y, Liu J, Hong Z, Gong M, Fan M, Han J, Liu J, Ding E, Wang J
  (2022) Few-shot font generation by learning fine-grained local styles. In:
  Proceedings of the IEEE/CVF Conference on Computer Vision and Pattern
  Recognition, pp 7895--7904

\bibitem[{Tian(2017)}]{Related22}
Tian Y (2017) zi2zi: Master chinese calligraphy with conditional adversarial
  networks. Internet] https://github com/kaonashi-tyc/zi2zi 3

\bibitem[{Wang et~al.(2004)Wang, Bovik, Sheikh, and Simoncelli}]{Related48}
Wang Z, Bovik AC, Sheikh HR, Simoncelli EP (2004) Image quality assessment:
  from error visibility to structural similarity. IEEE transactions on image
  processing 13(4):600--612

\bibitem[{Wen et~al.(2021)Wen, Li, Han, and Yuan}]{Related23}
Wen Q, Li S, Han B, Yuan Y (2021) Zigan: Fine-grained chinese calligraphy font
  generation via a few-shot style transfer approach. In: Proceedings of the
  29th ACM International Conference on Multimedia, pp 621--629

\bibitem[{Wolleb et~al.(2022)Wolleb, Sandk{\"u}hler, Bieder, and
  Cattin}]{Related21}
Wolleb J, Sandk{\"u}hler R, Bieder F, Cattin PC (2022) The swiss army knife for
  image-to-image translation: Multi-task diffusion models. arXiv preprint
  arXiv:220402641

\bibitem[{Xie et~al.(2021)Xie, Chen, Sun, and Lu}]{Related30}
Xie Y, Chen X, Sun L, Lu Y (2021) Dg-font: Deformable generative networks for
  unsupervised font generation. In: Proceedings of the IEEE/CVF Conference on
  Computer Vision and Pattern Recognition, pp 5130--5140

\bibitem[{Yang et~al.(2018)Yang, Xie, and Wang}]{Related09}
Yang X, Xie D, Wang X (2018) Crossing-domain generative adversarial networks
  for unsupervised multi-domain image-to-image translation. In: Proceedings of
  the 26th ACM international conference on Multimedia, pp 374--382

\bibitem[{Yi et~al.(2017)Yi, Zhang, Tan, and Gong}]{Related04}
Yi Z, Zhang H, Tan P, Gong M (2017) Dualgan: Unsupervised dual learning for
  image-to-image translation. In: Proceedings of the IEEE international
  conference on computer vision, pp 2849--2857

\bibitem[{Zeng et~al.(2021)Zeng, Chen, Liu, Wang, and Yao}]{Related25}
Zeng J, Chen Q, Liu Y, Wang M, Yao Y (2021) Strokegan: Reducing mode collapse
  in chinese font generation via stroke encoding. In: Proceedings of the AAAI
  Conference on Artificial Intelligence, pp 3270--3277

\bibitem[{Zhang and Agrawala(2023)}]{Related55}
Zhang L, Agrawala M (2023) Adding conditional control to text-to-image
  diffusion models. arXiv preprint arXiv:230205543

\bibitem[{Zhang et~al.(2018{\natexlab{a}})Zhang, Isola, Efros, Shechtman, and
  Wang}]{Related49}
Zhang R, Isola P, Efros AA, Shechtman E, Wang O (2018{\natexlab{a}}) The
  unreasonable effectiveness of deep features as a perceptual metric. In:
  Proceedings of the IEEE conference on computer vision and pattern
  recognition, pp 586--595

\bibitem[{Zhang et~al.(2018{\natexlab{b}})Zhang, Zhang, and Cai}]{Related46}
Zhang Y, Zhang Y, Cai W (2018{\natexlab{b}}) Separating style and content for
  generalized style transfer. In: Proceedings of the IEEE conference on
  computer vision and pattern recognition, pp 8447--8455

\bibitem[{Zhao et~al.(2022)Zhao, Bao, Li, and Zhu}]{Related19}
Zhao M, Bao F, Li C, Zhu J (2022) Egsde: Unpaired image-to-image translation
  via energy-guided stochastic differential equations. arXiv preprint
  arXiv:220706635

\bibitem[{Zhu et~al.(2017{\natexlab{a}})Zhu, Park, Isola, and
  Efros}]{Related03}
Zhu JY, Park T, Isola P, Efros AA (2017{\natexlab{a}}) Unpaired image-to-image
  translation using cycle-consistent adversarial networks. In: Proceedings of
  the IEEE international conference on computer vision, pp 2223--2232

\bibitem[{Zhu et~al.(2017{\natexlab{b}})Zhu, Zhang, Pathak, Darrell, Efros,
  Wang, and Shechtman}]{Related10}
Zhu JY, Zhang R, Pathak D, Darrell T, Efros AA, Wang O, Shechtman E
  (2017{\natexlab{b}}) Toward multimodal image-to-image translation. Advances
  in neural information processing systems 30

\end{thebibliography}


\end{document}